\def\year{2021}\relax
\documentclass[letterpaper]{article} 
\usepackage{aaai21}  
\usepackage{times}  
\usepackage{helvet} 
\usepackage{courier}  
\usepackage[hyphens]{url}  
\usepackage{graphicx} 
\usepackage{natbib}  
\usepackage{caption} 
\usepackage{booktabs}
\usepackage{pgfplots}
\usepackage{pgfplotstable}
\pgfplotsset{compat=1.7}
\usepackage{tikz}
\usepackage{appendix}

\title{Language Model Priming for Cross-Lingual Event Extraction}
\author {
    Steven Fincke, Shantanu Agarwal, Scott Miller, Elizabeth Boschee \\
    
}

\affiliations{
    University of Southern California Information Sciences Institute \\
    sfincke@isi.edu, shantanu@isi.edu, smiller@isi.edu, boschee@isi.edu
}
\begin{document}

\maketitle

\begin{abstract}
We present a novel, language-agnostic approach to ``priming'' language models for the task of event extraction, providing particularly effective performance in low-resource and zero-shot cross-lingual settings. 
With priming, we augment the input to the transformer stack's language model differently depending on the question(s) being asked of the model at runtime. 
For instance, if the model is being asked to identify arguments for the trigger \textit{protested}, we will provide that trigger as part of the input to the language model, allowing it to produce different representations for candidate arguments than when it is asked about arguments for the trigger \textit{arrest} elsewhere in the same sentence.
We show that by enabling the language model to better compensate for the deficits of sparse and noisy training data, our approach improves both trigger and argument detection and classification significantly over the state of the art in a zero-shot cross-lingual setting.

\end{abstract}

\section{Introduction}

Recent advances in massively-pretrained cross-lingual language models, e.g.\ \citet{conneau-etal-2020-unsupervised}, have revolutionized approaches to extracting information from a much broader set of languages than was previously possible. For some information extraction (IE) tasks, it is not unusual to find annotated datasets in a variety of languages. Name annotation, for instance, has both wide utility and can be performed relatively cheaply and easily by non-experts---and as a result, one can find annotated named entity datasets for languages from Spanish to Polish to Farsi to Indonesian. In contrast, datasets for event extraction tasks are much fewer and further between (even in English). 

In order to extract events from most languages, then, it is critical to be able to train models with data from one language and deploy those models in another. A pretrained, cross-lingual language model can carry some of this burden. If the language model's vector representation for \textit{arrest} (English) is close to its representation for \textit{anholdelsen} (Danish), then a model trained on the English sentence ``\textit{They protested his arrest}'' may be able to detect the \textsc{Arrest-Jail} event in ``\textit{De tre m{\ae}nd blev ved \textbf{anholdelsen} 29 januar}"\footnote{The three men were arrested on January 29. (Danish)}.

One major advance in recent years was the move from static word embeddings, e.g. word2vec \cite{NIPS2013_9aa42b31}, where the representation for a word is constant no matter the context in which it appears, to contextualized language models, e.g. ELMo \cite{peters-etal-2018-deep} and BERT \cite{devlin-etal-2019-bert}, where the representation for a word is conditioned on its surrounding context. With contextualized language models, the representation for \textit{arrest} in the context "\textit{They protested his arrest}" will likely be very different than when it appears in "\textit{She is a cardiac arrest survivor}". 

However, even with a powerful mechanism like BERT, the representation for, say, \textit{activists}, in ``\textit{Activists protested his arrest}'' will be the same whether the model is being asked if \textit{activists} is an argument of the \textsc{Demonstrate} event \textit{\textbf{protested}} or of the \textsc{Arrest-Jail} event \textit{\textbf{arrest}}. 

Recent advances for structured NLP tasks, such as named entity recognition \cite{li2019unified}, relation extraction \cite{li-etal-2019-entity} and coreference resolution \cite{wu-etal-2020-corefqa}, have noted this and responded accordingly, by providing a \textit{prompt} that modifies the context in which a sentence is seen, allowing the language model to adjust its representations of the sentence tokens accordingly. This method for injecting task-specific guidance to the model is particularly beneficial in low-resource settings \cite{DBLP:journals/corr/abs-2103-08493}.

Returning to the task of event extraction, \citet{du2020event} show some of the promise of applying similar principles in this domain, reframing event extraction as a question-answering problem. Their best results come from using natural English questions as prompts to their overall model, e.g. asking ``\textit{Which is the \underline{agent} in \underline{arrested}? | The police arrested the thief.}'' However, it is not clear that this English-centric approach of question generation will generalize well to cross-lingual transfer.

In this work, we present a new, language-agnostic mechanism \textsc{IE-Prime} which provides task-specific direction to an event extraction model at runtime. We show that it significantly outperforms prior work on argument extraction, including \citet{du2020event}, and that it is particularly effective in low-resource and zero-shot cross-lingual settings for both trigger and argument detection and classification. 

\section{Related Work}\label{sec:relatedwork}

Event extraction is a well-studied topic. Some of the most effective recent approaches in a mono-lingual context include \citet{wadden-etal-2019-entity} (DyGIE++), which combines local and global context using contextualized span representations across three unified subtasks (name, relation, and event extraction), and \citet{lin-etal-2020-joint} (OneIE), which uses a joint neural framework to extract globally optimal IE results as graphs from input sentences. Our specific approach draws inspiration from work in question answering. Although several recent works have been likewise inspired, (e.g. \citet{du2020event}, \citet{feng2020probing}, \cite{liu-etal-2020-event}), many of these approaches use natural-language questions and are not therefore an ideal fit for cross-lingual applications. Our system is also trained to inherently extract multiple ``answers'' from a given question context, e.g.\ to simultaneously identify more than one \textsc{Victim} of an \textsc{Attack} event. This is an important improvement over previous QA-inspired event extraction frameworks, which trained models to extract only single answer spans and would extract multiple spans at decode time by making selections based on a probability cutoff; our approach avoids this discrepancy between training and decoding behavior.
We compare against relevant prior work in both monolingual and cross-lingual contexts and show improved performance for argument extraction in all conditions and improved performance for trigger extraction in the cross-lingual context. 

The primary focus of this work is on cross-lingual event extraction. However, most recent cross-lingual event extraction work (e.g.\ \citet{subburathinam-etal-2019-cross}, \citet{ahmad2020gate}, and \citet{nguyen-nguyen-2021-improving}) focuses solely on the event argument role labeling task (ignoring trigger detection) and requires pre-generated entity mentions to serve as candidate arguments. Moreover, all of these approaches rely on structural features in both languages, including but not limited to dependency parses. In contrast, our approach does not require any extraction of linguistic structure in the target language. Our argument extraction approach also addresses the problem of candidate identification in addition to labeling, which is critical in real-world applications. We report results both on the limited task performed by the above-mentioned papers, showing improvements over the reported state of the art, as well as on the complete end-to-end task.

\section{Approach}

In this section we describe our baseline system (\textsc{IE-Baseline}) and an extension of that system (\textsc{IE-Prime}) that provides significant improvement. Both systems are designed to be completely language-agnostic---documents in any language supported by the language model can be used as either training or test data---and as such can be applied in either a monolingual or cross-lingual context. 

\subsection{\textsc{IE-Baseline}}
\label{sec:baseline}

Our baseline system consists of two trained components, one for event trigger extraction and one for argument attachment. Trigger detection and classification is performed using a simple beginning-inside-outside (BIO) sequence-labeling architecture composed of a single linear classification layer on top of the transformer stack.

\begin{figure}[t]
\centering
\includegraphics[width=0.9\columnwidth]{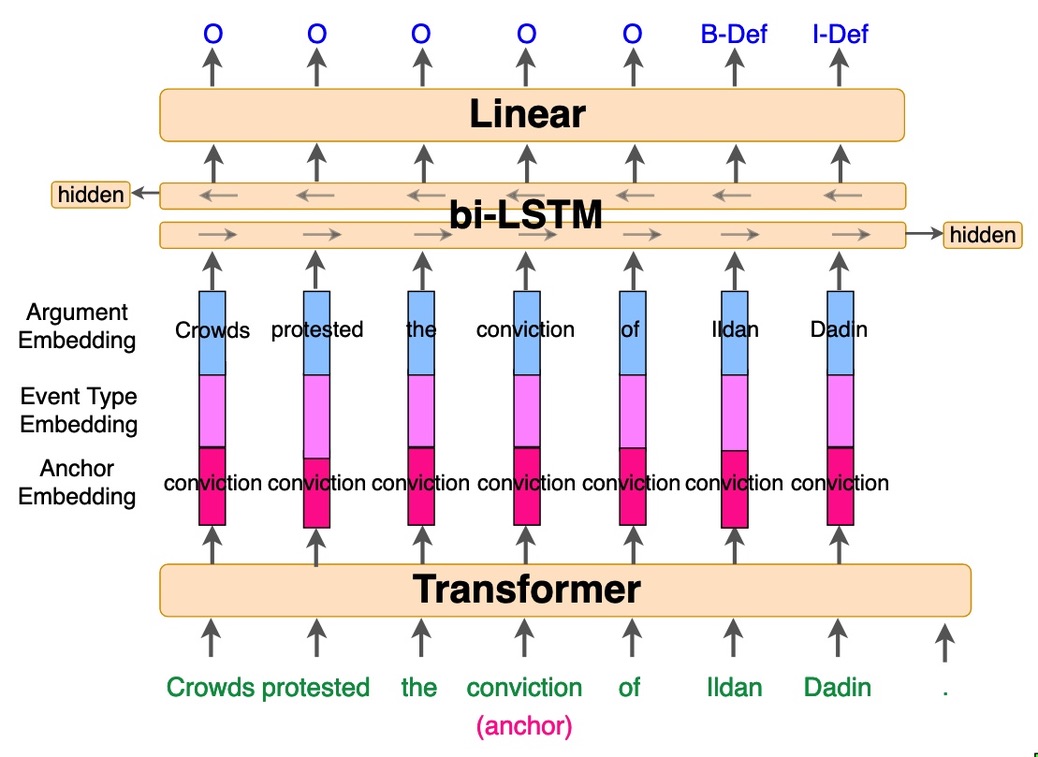}
\caption{Baseline argument attachment architecture.}
\label{fig:baseline-aa}
\end{figure}

Our baseline argument extraction system (shown in Figure \ref{fig:baseline-aa}) takes as its input a proposed event trigger and from an input sentence identifies argument spans and labels them with argument roles. 
This system produces arguments by generating BIO argument labels over the sentence tokens using a bi-LSTM, feeding into a single linear layer and then to a CRF-based loss function\footnote{As implemented in torchcrf, https://pytorch-crf.readthedocs.io/en/stable/} atop a transformer-based language model. The input for each token is the concatenation of the transformer output for the token itself, along with the transformer output for the trigger token, and an embedding for the event type; in our default configuration, no input entity information is used. When an individual input token is split up by the language model's tokenizer (e.g.\ if \textit{characteristically} is split into \textit{characteristic} and \textit{\#\#ally}), the average of the output vectors for the parts represents the whole.\footnote{We also explored other strategies, e.g.\ taking only the first vector, but averaging worked best.} All models fine tune all the layers of the language model and only use the output from the final layer. 

\subsection{\textsc{IE-Prime}}

The key to our proposed system, \textsc{IE-Prime}, is ``priming'': the idea that we can augment the input to the transformer stack in a way that provides critical additional information to the model at runtime.

We first describe how we do this for the task of \textit{argument extraction}. Our baseline argument extraction system is designed to take as input a single trigger and a sentence. It then produces a complete set of argument spans and roles with respect to that trigger. It repeats this process for every proposed event trigger.

The first form of priming we explore leverages that trigger. Specifically, we augment the input to the language model by pre-pending the trigger to the sentence being considered (divided from the sentence by a sentence-separating token appropriate to the language model being used, e.g.\ \texttt{[SEP]} for BERT). So, if a model trained using BERT is seeking arguments of \textit{protested}, the language model would receive the following input: 
{\small \texttt{[CLS] protested [SEP] crowds protested the conviction of Ildar Dadin [SEP]}}.

\begin{figure*}[t]
\centering
\includegraphics[width=11cm]{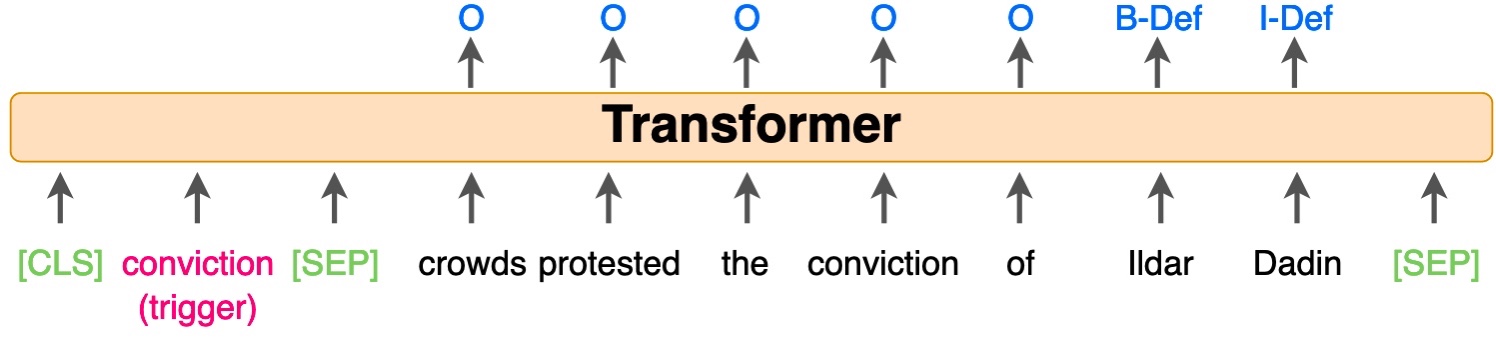}
\caption{Priming a sentence for the trigger \textit{conviction}. The span \textit{Ildar Dadin} is identified as a \textsc{Defendant} argument.}
\label{fig:triggerpriming}
\medskip
\centering
\includegraphics[width=13cm]{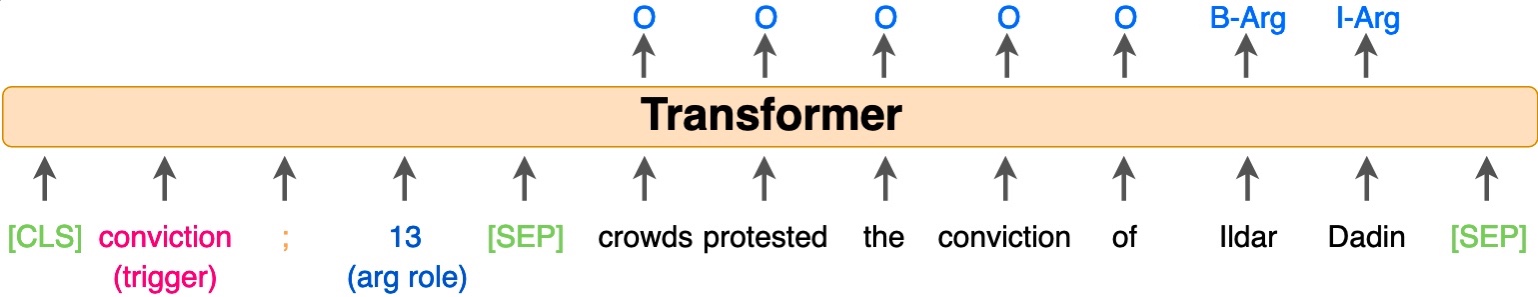}
\caption{Priming a sentence for the trigger \textit{conviction} and the argument role \textsc{Defendant}. The span \textit{Ildar Dadin} is identified as an argument and is therefore assigned the role being queried (\textsc{Defendant}).}
\label{fig:triggerandrolepriming}
\medskip
\centering
\includegraphics[width=11cm]{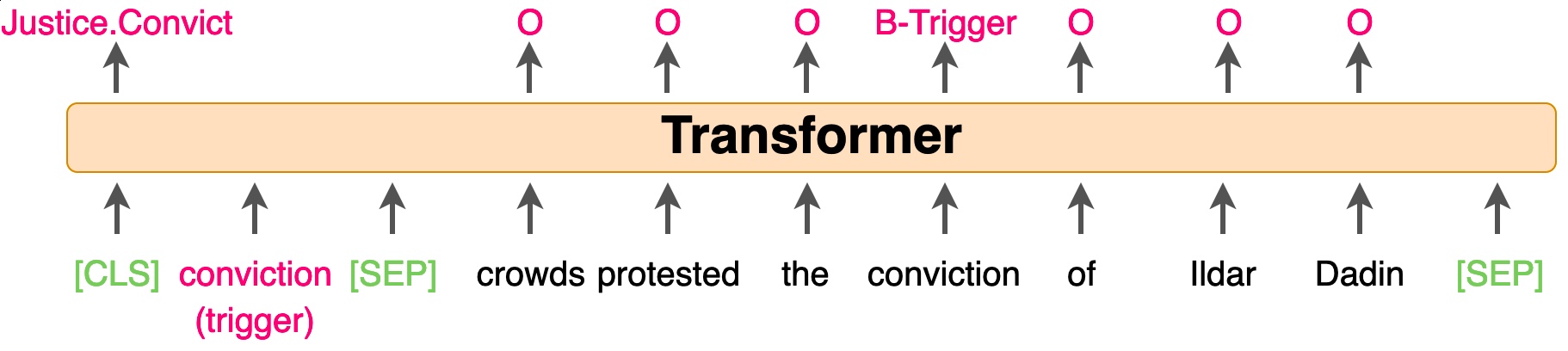}
\caption{Priming a sentence to determine whether the span \textit{conviction} is a trigger of type \textsc{Justice.Convict}.}
\label{fig:primedtrigger}
\end{figure*}

However, when searching for arguments of \textit{conviction}, the following input is used instead:
{\small \texttt{[CLS] conviction [SEP] crowds protested the conviction of Ildar Dadin [SEP]}}.
Figure \ref{fig:triggerpriming} shows a graphical representation of the second example. 

We note that our baseline system already indicates the trigger of interest when predicting arguments (by appending its vector to the vector representation for each token), so priming is arguably redundant. However, providing this input in a second way appears to enable the language model to respond more effectively to the change in focus at runtime. 

The second form of priming we explore leverages both the trigger and the argument role.\footnote{Different argument roles are possible for different ACE event types: our system only queries the roles allowed for the event type specified for the trigger.} Here, rather than asking the model for all arguments of a trigger at the same time, we query separately for each possible type of argument. That is, we query first for any \textsc{Defendant} arguments of a \textsc{Convict} event, and then separately for any \textsc{Adjudicator} arguments, and so on. 
In this formulation, we augment the input to the language model with both the trigger and the argument role. However, to facilitate application to a variety of languages, we replace each English argument role label with a unique integer string, giving a result like this when asking about any \textsc{Defendant} arguments of \textit{conviction}: 
{\small \texttt{[CLS] conviction ; 13 [SEP] crowds protested the conviction of Ildar Dadin [SEP]}}.
Figure \ref{fig:triggerandrolepriming} shows a graphical representation. 

Both forms of priming described above can be used to either generate BIO labels or label pre-identified candidate arguments. We found priming by trigger and argument role to be generally more successful and is what we report below unless otherwise noted.

Finally, we also developed a priming model for trigger detection and classification. Here, the input is primed with a single token from the sentence. So, the language model again sees something like this:
{\small \texttt{[CLS] conviction [SEP] crowds protested the conviction of Ildar Dadin [SEP]}}.
However, the question asked of the model is not about the arguments of \textit{conviction} but about whether \textit{conviction} is itself part of a trigger, and if so, what kind. This model targets two training objectives. One objective predicts the output span for the trigger with BIO labelling with a bi-LSTM and CRF, similar to argument extraction.
The other objective takes the concatenation of the language model output for the trigger token and the class token as input and applies a linear transformation to predict the event type.

Thus, in the example above, we expect \textit{conviction} to be marked as the trigger span and the language model outputs from \texttt{[CLS]} and \textit{conviction} to lead the model to generate \textsc{Justice.Convict} as the event type; figure \ref{fig:primedtrigger} shows this in graphical form. The system outputs a trigger only if and only if an event type is predicted; if BIO labelling provides no span overlapping with the priming token, the model outputs a trigger with just the priming token as its span.

\subsection{Variant: Pre-generated Candidate Arguments}
To allow better comparison with prior cross-lingual work (e.g. \citet{subburathinam-etal-2019-cross}), we developed a variant of our system which takes as input pre-generated candidate arguments instead of finding argument spans in the sentence at decode time. In this paper, we use this variant to produce results in an experimental setting that uses gold entity mentions as argument candidates. We adapt our architecture for this setting in three small ways. 

\textit{First}, it no longer makes sense to have the model consider arbitrary spans, so instead of all tokens, we present a sequence of candidate arguments and ask the model to classify each with respect to a specified trigger. (We re-use the same architecture: a single linear layer with a CRF-based loss function on top of the transformer stack.) For each candidate, we take as its representation the transformer output for the candidate's ``most representative'' token, ideally its linguistic head. For instance, the model will use the vector for \textit{Smith} to represent \textit{Bob Smith}, or the vector for \textit{student} to represent \textit{the medical student}. To select this token for each candidate argument, we take the highest-ranking token in its dependency parse, if available\footnote{Generated here using spaCy (https://spacy.io/) in English or the model trained from the Prague Arabic Dependency Treebank (https://github.com/UniversalDependencies/UD\_Arabic-PADT) for UDPipe \cite{straka-strakova-2017-tokenizing} in Arabic.}; if not, we simply select an argument's first token. (Using dependency parses is not required and typically provides only a small (1-2 point) gain over always selecting the first token.)

\textit{Second}, we augment our argument model to consider entity type by including the output vector from an entity type embedding in the set of vectors concatenated for the candidate.

\textit{Third}, we constrain our system to produce only ``legal'' argument/trigger pairs; that is, if no entity of type \textsc{Person} ever appears as the \textsc{Place} argument to an \textsc{Arrest-Jail} event in training, neither do we allow it to do so at test time.

We note that this approach could also be applied to system-generated candidate arguments, should training data for them exist (e.g.\ in the ACE2005 dataset). However, in our cross-lingual experiments, our primary system (which considers all possible spans as arguments) provided superior results, so we did not pursue this variant further except when required for experiments using gold entity mentions.

\begin{table}[]
\begin{tabular}{@{}llll@{}}
\toprule
                    & \multicolumn{2}{c}{Mono} & Cross       \\
                    \cmidrule(lr){2-3}\cmidrule(lr){4-4}
                    & en$\rightarrow$en & ar$\rightarrow$ar & en$\rightarrow$ar \\ \midrule
\citet{du2020event} & 65.4          &   --          &   --               \\
\citet{lin-etal-2020-joint} & 69.3 & -- & -- \\ \midrule
\textsc{IE-Baseline}         & 63.2          & 53.3          & 44.7                \\
\textsc{IE-Prime}            & \textbf{72.4} & \textbf{67.7} & \textbf{50.3}      \\ \bottomrule
\end{tabular}
\caption{Gains in argument classification F1 score from priming for argument extraction (using gold triggers in our primary experimental setting). As for all reported results in this paper, \textsc{IE-Prime} uses the trigger+role configuration for priming unless otherwise specified.}
\label{tab:priming}
\end{table}

\begin{table*}[t]
\centering
\begin{tabular}{@{}lllll@{}}
\toprule
                                      & Priming method & Recall & Precision & F-Measure           \\ \midrule
\citet{subburathinam-etal-2019-cross} & --& & & 61.8                \\
\citet{ahmad2020gate}                 & --& & & 68.5                \\ \midrule
\textsc{IE-Baseline}                  & --& 67.1 & 79.6 & 72.8      \\
\textsc{IE-Prime}                     & trigger + role & 66.5 & 82.9 & 73.8 \\ 
\textsc{IE-Prime}                     & trigger & \textbf{67.5} & \textbf{83.7} & \textbf{74.7} \\ \bottomrule
\end{tabular}
\caption{Secondary experimental setting: Argument classification F1 in the zero-shot cross-lingual condition (train on English, test on Arabic) with gold triggers and gold entity mentions, following splits from \cite{subburathinam-etal-2019-cross}.}
\label{tab:rpisplits}
\end{table*}

\begin{table*}[t]
\centering
\begin{tabular}{@{}lllllll@{}}
\toprule
& \multicolumn{4}{l}{Monolingal}    & \multicolumn{2}{l}{Cross-lingual}       \\ \midrule
& \multicolumn{2}{l}{en$\rightarrow$en} & \multicolumn{2}{l}{ar$\rightarrow$ar} & \multicolumn{2}{l}{en$\rightarrow$ar} \\
\cmidrule(lr){2-3}\cmidrule(lr){4-5}\cmidrule(lr){6-7}
& trigger & argument & trigger & argument & trigger & argument \\
\citet{wadden-etal-2019-entity}  & 69.7 & 48.8 & -- & --  & -- & -- \\
\citet{wadden-etal-2019-entity}$\dagger$ & 70.4 & 52.2 & \textbf{61.5} & 44.4  & 41.6 & 22.0 \\
\citet{du2020event}     & 72.4 & 53.1 & -- & -- & -- & -- \\ 
\citet{lin-etal-2020-joint}     & \textbf{74.7} & \textbf{56.8} & -- & -- & -- & -- \\ \midrule
\textsc{IE-Prime} (arguments only)                   & 71.2 & 55.3 & 61.2 & \textbf{48.9} & 42.4 & 30.2 \\
\textsc{IE-Prime} (arguments + triggers)           & 68.1 & 52.9 & 60.2 & 48.7 & \textbf{51.0} & \textbf{32.4} \\ \bottomrule
\end{tabular}
\caption{Trigger and argument classification F1 for end-to-end systems in our primary experimental setting. The first version of \textsc{IE-Prime} includes the baseline trigger component and the primed argument extraction component. The second version includes both the primed trigger component and the primed argument extraction component. $\dagger$ indicates our local re-run of \cite{wadden-etal-2019-entity}.}
\label{tab:endtoend}
\end{table*}

\section{Experimental Setup}

We report results in \textbf{two experimental settings}, both using the ACE 2005 corpus (English and Arabic)\footnote{https://www.ldc.upenn.edu/collaborations/past-projects/ace}. We believe the first setting provides a more accurate and complete picture of the full event extraction task, but we include the second (which evaluates only event-argument role labeling) for a full comparison to prior work.

Our \textbf{primary experimental setting} uses the standard English document train/dev/test splits for this dataset \cite{yang-mitchell-2016-joint} and the Arabic splits proposed by \citet{xu-etal-2021-gradual}. In both cases, we draw sentence breaks from the data made available by \citet{xu-etal-2021-gradual}, which is generated using the DyGIE++ codebase.\footnote{https://github.com/dwadden/dygiepp} We further split Arabic sentences at runtime to ensure a maximum length of 80 tokens (we still score against the unsplit reference). We also use Farasa \cite{abdelali-etal-2016-farasa} to remove tatweels and map presentation forms to the standard code range. All results reported in this paper other than Table \ref{tab:rpisplits} use this experimental setting and are the average of five seeds.

Our \textbf{secondary experimental setting} evaluates only event-argument role labeling. It replicates the conditions proposed by \citet{subburathinam-etal-2019-cross}, which ask systems to label individual instances consisting of a gold trigger and a gold entity mention. The train/dev/test split does not consider the origin of each instance, meaning that the same sentence and trigger can be found in both train and test, with different candidate arguments. This is inconsequential in a cross-lingual setting but should be taken into consideration if assessing monolingual results. Moreover, following prior work, both training and test down-sample the number of negative instances to match the number of positive instances. For this reason, observed precision is dramatically over-inflated compared to a real-world scenario; reported F-measure should be considered in this light. The only table in this paper using this experimental setting is Table \ref{tab:rpisplits}.

Unless otherwise specified, for \textbf{language models} we use the large, cased version of BERT \cite{devlin-etal-2019-bert} for the monolingual English condition and the large version of XLM-RoBERTa~\cite{conneau-etal-2020-unsupervised} for cross-lingual or Arabic-only conditions. We use BERT for monolingual English solely to ensure a fair comparison to prior work; XLM-RoBERTa could also be used with similar results.

Following community practice for \textbf{evaluation metrics}, e.g.\ \citet{Zhang2019JointEA}, we consider a trigger correct if its offsets and event type are correct, and we consider an argument correct if its offsets, event type, and role find a match in the ground truth.

\section{Results}

\subsection{Argument Extraction}
We begin by assessing the impact of our new priming architecture directly on argument extraction. To do this, we first present results with gold triggers in our primary experimental setting. Table \ref{tab:priming} shows improvements from priming across the board in both monolingual and cross-lingual conditions. We also report English results here from \citet{du2020event}, also inspired by a question-answering approach, and from OneIE \cite{lin-etal-2020-joint}\footnote{We train the full OneIE model using the released code and then constrain it to gold triggers at test time.}. \textsc{IE-Prime} comfortably out-performs both prior baselines in the monolingual English condition. 

In order to compare against prior work in a cross-lingual setting, we next present analogous results in our secondary experimental setting.\footnote{We are not aware of any published full system results for English$\rightarrow$Arabic cross-lingual ACE event extraction.} This setting represents the narrower task of event argument role labeling (using gold triggers and gold entity mentions). Table \ref{tab:rpisplits} presents our results in this experimental condition, where \textsc{IE-Prime} shows more than a six-point improvement over the best previously-reported results \cite{ahmad2020gate}. 

We note that this secondary experimental condition is somewhat artificial, not just in its reliance on gold entity mentions (both spans and entity types) but also in its exclusion of 90\% of the negative instances during both training \textit{and} test. 
To mitigate the significant dataset bias in this condition, we ran a version of our system which took as its output the union of arguments found by models trained with five different seeds, allowing us to force the system to prioritize recall. 

We also show results for both priming variants for the secondary experimental condition (priming by trigger only and by trigger and role together). In the primary experimental condition, the trigger+role approach to priming is always better and that is what we typically report in this paper (e.g.\ in Table \ref{tab:priming}). However, here, the trigger-only version is superior, perhaps because it allows the full model to see at once how many arguments are being predicted for a given trigger and therefore again allows the model to favor higher recall to better approximate the (somewhat unrealistic) distribution in this version of the data. 

\subsection{End-to-End System}

To further situate our system within the state of the art, we must consider an end-to-end system, since the currently best-performing event extraction systems jointly optimize extraction of entity mentions, triggers and arguments. Our primed system (as analyzed so far) focuses solely on improving argument extraction performance, and indeed our simple model's trigger performance lags behind the state of the art, particularly in English. Our system also does not use entity annotation, which is used by both of the prior-work baselines shown here. Still, we see in Table \ref{tab:endtoend} that despite weaker trigger performance and without using any entity annotation, our monolingual argument extraction in English comes near that of the state-of-the-art system.\footnote{Note that since this is an end-to-end system, trigger performance directly impacts argument extraction performance, as systems are penalized for producing arguments for spurious triggers or missing arguments for missed triggers.}

Arabic event extraction performance has not been as widely reported as English, and the Arabic training set is only 40\% of the size of the English training set, making it a lower-resource condition. Following \citet{xu-etal-2021-gradual}, we take as a baseline a DyGIE++ model trained on the Arabic dataset. Results in Arabic (monolingual or cross-lingual) confirm the strength of our argument extraction approach with respect to the state of the art, both in the lower-resource Arabic monolingual condition (+4.5 over DyGIE++) and the zero-shot cross-lingual condition (+8.2 over DyGIE++).

Finally, noting that triggers serve as a point of weakness in our baseline system, we further explore the possibilities of priming by testing our primed trigger model in the cross-lingual context. Although we see no gain in the monolingual conditions, we see a very significant gain in the cross-lingual context (from 42.4 to 51.0 for trigger classification), suggesting that the primary strength of priming is enabling the underlying language model to compensate when the task-specific training data is noisy (e.g.\ from another language).

\begin{figure}[t]
\centering
\includegraphics[width=1.0\columnwidth]{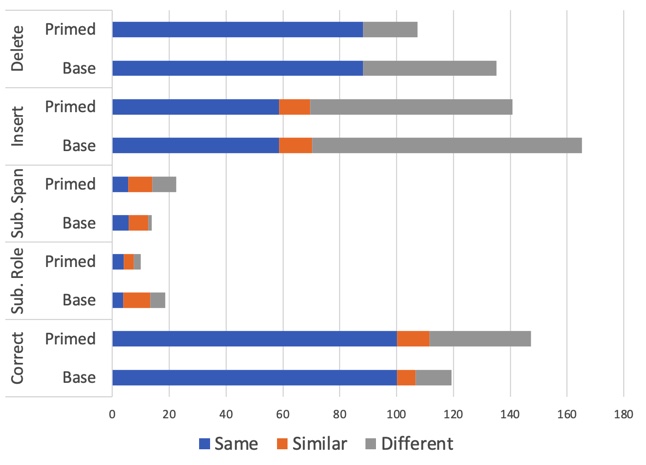}
\caption{Analysis of differences between \textsc{IE-Baseline} and \textsc{IE-Prime} for zero-shot Arabic with gold triggers (averaged over five seeds). } \
\label{fig:base_prime_diff}
\end{figure}

\begin{figure}[t]
\centering
\includegraphics[width=1.0\columnwidth]{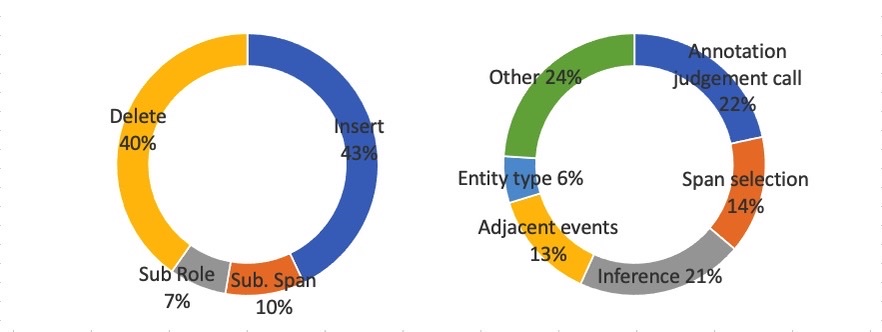}
\caption{Classification of errors when training and testing on English with gold triggers.}
\label{fig:error-donughts}
\end{figure}

\section{Further Analysis}
\paragraph{Error Analysis}
To focus on the impact of priming, we first analyze differences between the output of \textsc{IE-Baseline} and \textsc{IE-Prime} in the zero-shot English$\rightarrow$Arabic context (Figure \ref{fig:base_prime_diff}). Cases in which a system output span overlaps with with the gold but the exact span or role are incorrect are classified as substitutions; instances when both systems provide outputs which overlap but differ in exact span or role are marked \textit{similar}.
Results show that \textsc{IE-Prime} provides significantly more correct answers overall, adding an additional 47 beyond the 100 found by both systems, while \textsc{IE-Baseline} adds only an additional 19 not found by \textsc{IE-Prime}. The total number of substitutions remains constant, and the increase in correct guesses for \textsc{IE-Prime} yields a 60\% reduction in novel deletions (misses). Comparing deletions to insertions, the two systems are more likely to agree on deletions (items missed by the model), while insertions are much more heterogeneous. This is intuitive: there are more unique ways to hallucinate a false alarm argument than there are unique correct answers to delete. We note that \textsc{IE-Prime} does reduce novel insertions by a quarter compared to \textsc{IE-Baseline}, showing that the improvement from priming holds for both recall and precision.

\begin{table}[t]
\centering
\setlength\tabcolsep{5pt}
\begin{tabular}{@{}lllllll@{}}
\toprule
            & \multicolumn{2}{c}{en$\rightarrow$en} & \multicolumn{2}{c}{ar$\rightarrow$ar} & \multicolumn{2}{c}{en$\rightarrow$ar} \\ 
            
            \cmidrule(lr){2-3}\cmidrule(lr){4-5}\cmidrule(lr){6-7}
            & base & large & base & large  & base & large \\\midrule
\textsc{IE-Baseline} & 60.0 & 66.0           & 46.9  & 53.3           & 35.6 & 44.7   \\
\textsc{IE-Prime}    & 69.0 & \textbf{74.1}  & 60.8 & \textbf{67.7}   & 40.2 & \textbf{50.3}  \\\bottomrule
\end{tabular}
\caption{Comparison of argument classification F1 (using gold triggers) based on size of pretrained language model.}
\label{tab:pretrained}
\end{table}

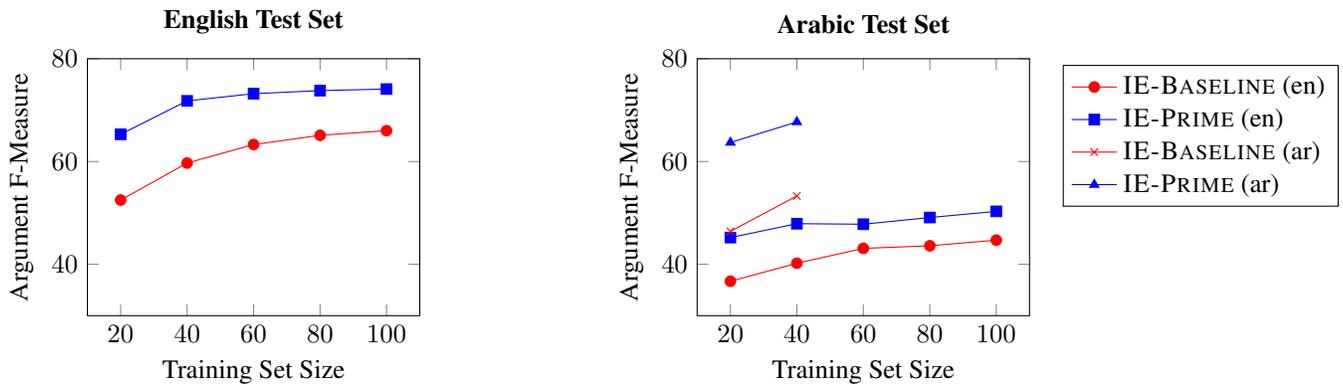
\begin{figure*}[t]
\centering
\begin{tikzpicture}
\begin{axis}[
    title=\textbf{English Test Set},
	xlabel=Training Set Size,
	ylabel=Argument F-Measure,
	ymin=30,
	ymax=80,
	xmin=10,
	xmax=110,
	width=6cm,height=5cm
    ]
\addplot[color=red,mark=*] coordinates {
(20, 52.5)
(40, 59.7)
(60, 63.3)
(80, 65.1)
(100, 66.0)
};
\addplot[color=blue,mark=square*] coordinates {
(20, 65.3)
(40, 71.8)
(60, 73.2)
(80, 73.8)
(100, 74.1)
};
\end{axis}
\end{tikzpicture}
\hfill
\begin{tikzpicture}
\begin{axis}[
    title=\textbf{Arabic Test Set},
	xlabel=Training Set Size,
	ylabel=Argument F-Measure,
	ymin=30,
	ymax=80,
	xmin=10,
	xmax=110,
	width=6cm,height=5cm,
	legend cell align={left},
    legend style={at={(1.1,0.7)},anchor=west}
    ]
\addplot[color=red,mark=*] coordinates {
(20, 36.7)
(40, 40.2)
(60, 43.1)
(80, 43.6)
(100, 44.7)
};
\addplot[color=blue,mark=square*] coordinates {
(20, 45.2)
(40, 47.9)
(60, 47.8)
(80, 49.1)
(100, 50.3)
};
\addplot[color=red,mark=x] coordinates {
(20, 46.4)
(40, 53.3)
};
\addplot[color=blue,mark=triangle*] coordinates {
(20, 63.7)
(40, 67.7)
};
\legend{\textsc{IE-Baseline} (en), \textsc{IE-Prime} (en), \textsc{IE-Baseline} (ar), \textsc{IE-Prime} (ar)}
\end{axis}
\end{tikzpicture}
\caption{Comparison of \textsc{IE-Baseline} and \textsc{IE-Prime} by approximate training set size. Training size here is calculated as the number of events in a document set and is shown as a percentage of full English training set size. The language of the data in which the models were trained are denoted in parenthesis. Experiments in this figure use gold triggers.}
\label{fig:trainingsetsize}
\end{figure*}

For qualitative analysis, we focus on \textsc{IE-Prime} in the monolingual English condition.\footnote{Only the system using the seed \textit{1235} was examined.} Figure \ref{fig:error-donughts} summarizes our observations. Overall, we find the system evenly balanced between deletions (40\% of all errors) and insertions (43\%). Another 17\% are classified as substitutions, with only one of the role (7\%) or argument span (10\%) being incorrect. We classify observed errors into six categories:

\begin{itemize}
    \item Annotation judgment call (22\%): The system decision seems plausibly correct to a new reader.
    \item Span selection errors (14\%): The model did not select the correct span for an argument, e.g. labelling \textit{Headquarters} when \textit{Security Headquarters} is expected. A higher-level re-scoring of argument spans might be helpful here.
    \item Inference required (21\%): Sometimes the system task requires world knowledge and/or knowledge of document context beyond the local sentence (which our model does not consider).
    For instance, an earlier sentence might give a clue that a person is a smuggler, making them more likely to be later found as a  \textsc{Defendant} in a \textsc{Convict} event; incorporating document context is an important next step for our approach.
    \item Confusion due to adjacent events (13\%): Some confusion can arise due to the proximity in the text of other events. For instance, in \textit{Davies is leaving to become chairman of the London School of Economics}, our model correctly marks \textit{London School of Economics} as the \textit{Entity} argument for \textit{become (chairman)}, but we incorrectly also label it as the \textit{Entity} for \textit{leaving}. Adding a mechanism to ensure consistency between events could help here.
    \item Entity type (6\%). A handful of errors involve entities whose semantic class is easy to mistake, e.g. a model likely not understanding \textit{Milton Keynes} as a place name.
    \item Other (24\%): The remaining errors lack any ready explanation, but it appears that some might benefit from more explicit modeling of syntactic information or a re-scoring pass to consider event-to-event interaction.

\end{itemize}
\paragraph{Pretrained Model Size}
We believe the gains from priming come from giving the base language model more opportunity to compensate for gaps in the training data. Another common way to improve the performance of an architecture that relies on a pretrained language model is simply to increase the size of that model (either the number of parameters users, or the data sets it sees during training, or both). Table \ref{tab:pretrained} presents the results of our baseline and primed models with both the base and large versions of XLM-RoBERTa. As we can see, the gains achieved from priming and from language model size are complementary. In all conditions, we see an improvement over baseline from adding \textit{either} priming or moving to a larger language model, and again in all conditions, we see significant further gains from adding both. We note that the absolute gains from adding priming are quite similar regardless of language model size.

\paragraph{Training Set Size}
We have hypothesized that priming is particularly effective in low-resource conditions. To test this, we vary the size of our training set and examine the relative gain from priming in each condition. For simplicity, Figure \ref{fig:trainingsetsize} presents results only for argument extraction, with gold triggers.
Training size is calculated as the number of events in a document set and shown as a percentage of the English training set size; so, the full Arabic training set (1.7K events) is approximately equivalent to 40\% of the English training set (4.2K events). 

We make two primary observations. First, the gap between the baseline and primed systems is indeed greatest in the lowest-resource settings, showing that the priming architecture provides more power when the model is under-resourced. For instance, when only 20\% of the English data is available ($\sim$850 events), we see a 12.8-point gain from adding priming; when 100\% of the data is available, the gain is smaller (8.1 points). The same holds true for the Arabic monolingual and English$\rightarrow$Arabic cross-lingual conditions.

Second, we show that priming is able to overcome low-resource conditions rather significantly. In English, \textsc{IE-Prime} requires only 20\% of the training to nearly equal the performance of \textsc{IE-Baseline} trained with 100\% of the data. The same is true in the cross-lingual condition. We also see that in the this lowest-resource condition (20\%, or $\sim$850 events), the performance of \textsc{IE-Prime} in the cross-lingual condition is actually almost equal to the performance of \textsc{IE-Baseline} trained on the same amount of native Arabic data. All of these results show the utility of priming when only a small amount of data (or data from the wrong language) is available.

\section{Conclusions \& Future Work}

We have shown here that our novel priming architecture improves both trigger and argument detection and classification significantly over the state of the art in a zero-shot cross-lingual setting. Our approach also provides significant gains for argument extraction in the monolingual context. However, there are still many areas yet to be explored within this new paradigm. For instance, prior work has shown improvement by jointly modeling triggers and arguments; we expect this would be an area of gain here as well. Prior work has also shown the value of document-level information, none of which is exploited in our current work. We look forward to investigating the incorporation of these elements (and others) as we continue to explore this promising paradigm.

\section*{Acknowledgements}
This research is based upon work supported in part by the Office of the Director of National Intelligence (ODNI), Intelligence Advanced Research Projects Activity (IARPA), via Contract No. 2019-19051600007. The views and conclusions contained herein are those of the authors and should not be interpreted as necessarily representing the official policies, either expressed or implied, of ODNI, IARPA, or the U.S. Government. The U.S. Government is authorized to reproduce and distribute reprints for governmental purposes notwithstanding any copyright annotation therein.

\end{document}